%% file: iclr2023_conference_tinypaper.tex
\documentclass{article} 
\usepackage{iclr2023_conference_tinypaper,times}

\input{math_commands.tex}

\usepackage{hyperref}
\usepackage{url}
\usepackage{amsthm}
\usepackage{amssymb}
\usepackage{booktabs}
\usepackage{adjustbox}

\title{Exploring Loss Design Techniques For Decision Tree Robustness To Label Noise}

\author{Lukasz Sztukiewicz\thanks{Preferred contact: research@lukaszsztukiewicz.com} \space , Jack H. Good, Artur W. Dubrawski \\
Auton Lab, Robotics Institute, Carnegie Mellon University \\
\texttt{\{lsztukie, jhgood, awd\}@andrew.cmu.edu} \\
}

\newcommand{\fix}{\marginpar{FIX}}
\newcommand{\new}{\marginpar{NEW}}

\newtheorem{theorem}{Theorem}

\iclrfinalcopy 
\begin{document}

\maketitle

\begin{abstract}
In the real world, data is often noisy, affecting not only the quality of features but also the accuracy of labels. Current research on mitigating label errors stems primarily from advances in deep learning, and a gap exists in exploring interpretable models, particularly those rooted in decision trees. In this study, we investigate whether ideas from deep learning loss design can be applied to improve the robustness of decision trees. In particular, we show that loss correction and symmetric losses, both standard approaches, are not effective. We argue that other directions need to be explored to improve the robustness of decision trees to label noise.
\end{abstract}

\textbf{INTRODUCTION AND RELATED WORK}

Label errors, also known as label noise, are widespread. 
Recent estimates indicate that commonly used research data, which is usually assumed to be clean, contains up to 10\% label corruption~\citep{northcutt2021labelerrors}. 
This problem has received considerable attention in the deep learning community, resulting in several techniques to improve robustness of neural networks to incorrect labels. 
These methods are surveyed in~\cite{han_survey_2021} and~\cite{9729424}.
Despite these efforts, work on interpretable machine learning methods, such as decision trees, remains limited. 
Existing research highlights the resilience of decision trees to symmetric label noise in specific scenarios, particularly in binary classification with large sample sizes \cite{ghosh_robustness_2017}. However, broader scenarios and enhancements in this context remain understudied. Most efforts involving tree-based algorithms focus on improving robustness of ensembles such as Random Forest~\citep{yang_robust_2019,8658395} or gradient boosting~\citep{7273923}, with little attention given to individual decision trees. 
This raises a natural question: Can recent advances in deep learning be applied to conventional decision tree induction?
 Recognizing the limited exploration of this research direction \cite{nanfack_constraint_2022}, our study aims to investigate this question.
In particular, we investigate whether loss design approaches including loss correction and symmetric loss functions can be effectively adapted to tree-based algorithms.


\textbf{THEORETICAL ANALYSIS}

We consider $c$-class classification problems. 
Let $\mathcal{X} \in \mathbb{R}^m$ be the feature space and $\mathcal{Y} \in \mathbb{R}^c$ be the label space. 
Let $(X,Y) \in \mathcal{X} \times \mathcal{Y}$ be random variables from which we observe the data set $\{(x_i,y_i)\}_{i=1}^{n}$, where each $x_i$ is a vector representing $m$ features and $y_i$ is a one-hot encoded vector label indicating one of $c$ classes.
The observed data are assumed to be affected by noise;
each $(x_i,y_i)$ is drawn from a distribution $(X,\tilde{Y})$
which is related to the true labels by the noise transition matrix
$T(x) \in [0,1]^{c \times c}$, where $T_{a,b}(x) = P(\Tilde{Y}=b|Y=a, X=x)$. 

A loss function $\mathcal{L}$ maps a true and predicted label
to a loss value;
learning searches for a model that minimizes the empirical loss
$\frac{1}{n}\sum_{i=1}^n \mathcal{L}(y_i,\hat y_i)$ for predictions $\hat y_i$.
Decision tree learning using Gini gain or information gain
minimizes the empirical mean squared error or cross-entropy loss, respectively, as shown in Appendix~\ref{sec:loss_equivalence}.
For both, the loss-minimizing values $v_\ell\in\mathbb{R}^c$ for leaves $\ell$ are the weighted means
$
v_\ell=\arg\min_v \sum_{i=1}^n \mu_\ell(x_i) \mathcal{L}(y_i,v)=\sum_{i=1}^n \mu_\ell(x_i)y_i/(\sum_i^n\mu_\ell(x_i))
$
where $\mu_\ell$ is the data membership function in leaf $\ell$, taking values in $\{0,1\}$.

\textbf{Impurity-based tree growth is invariant to forward loss correction.}
Loss correction \citep{8099723} assumes the transition matrix $T$ is known 
and incorporates it into the loss function
so that the loss minimizer is, in expectation,
the same as if the model had been trained on clean data.
In particular, \textit{forward correction} uses corrected loss
$\mathcal{L}_T(y_i,\hat{y}^i) = \mathcal{L}(y_i,T\hat{y}^i)$.
\begin{theorem}
\label{thm:forward_correction}
For any loss function where the minimizing leaf value is the weighted mean,
the loss value for a given tree structure is invariant to forward loss correction.
\end{theorem}
The proof is in Appendix~\ref{sec:proof_forward}.
From this we conclude that, while forward loss correction may change leaf values,
it ultimately does not affect the learned tree structure.
Therefore forward correction can be applied by simply learning a tree as usual,
then replacing the leaf values with $v_\ell\gets T^{-1}v_\ell$,
as shown in the proof.
Typically, however, the true class is observed the most frequently,
so this correction is unlikely to change the plurality class of leaves,
and thus unlikely to improve performance.

In addition to forward correction,
there is a related method called \textit{backward correction} \citep{8099723}.
For decision trees, backward correction reduces to simply fitting to data $\{(x_i,T^{-1}y_i)\}_{i=1}^n$, as shown in the Appendix~\ref{sec:proof_backward}.
Unlike forward correction, this can result in different tree structure.

\textbf{Symmetric loss functions are not suitable for decision tree growth.}
Symmetric loss functions have the property that there exists some constant $C$ such that, for any $\hat{y}$, $\sum_{k=1}^c \mathcal{L}(e_k,\hat{y}) = C$, where $e_k$ is the indicator (one-hot) vector with 1 at index $k$. 
Under certain assumptions, they are theoretically tolerant to label noise
and improve performance of models trained on data with noisy labels
compared to models trained with conventional loss functions~\citep{pmlr-v97-charoenphakdee19a}.
Here we assume one-hot training labels, that is, 
that there are no pseudo-labels in the data.
\begin{theorem}
\label{thm:symmetric_losses}
For any symmetric, non-negative loss function,
there exist loss-minimizing leaf values that are plurality indicators.
\end{theorem}
The proof is in Appendix~\ref{sec:proof_symmetric}.
Here ``plurality'' is a multi-class generalization of majority that refers to the most frequent class, without implying that its frequency is greater than one half. 
The assignment of leaf values to a plurality class indicator poses challenges to tree growth because, particularly with imbalanced data, it frequently occurs that all potential splits lead to both children nodes having the same leaf value, resulting in zero gain. Thus, symmetric loss functions are not suitable for decision tree growth.

\textbf{EMPIRICAL ANALYSIS}

We evaluate the impact of these types of loss correction on the performance of decision tree-based models.
We include decision trees, random forests, and ExtraTrees using implementations from the popular scikit-learn library.
We use six benchmark data sets from the OpenML data set repository~\citep{vanschoren_openml_2014},
outlined in Appendix Table~\ref{tab:datasets}.
We add a type of label noise called Noise Completely At Random ~\citep{Frnay2014ACI}, 
where each sample has a probability of $\eta$ to be flipped to another label uniformly at random.
We use $\eta$ from 0\% to 40\% in increments of 10\%.

\begin{figure}[h!]
\centering
\includegraphics[scale=0.45]{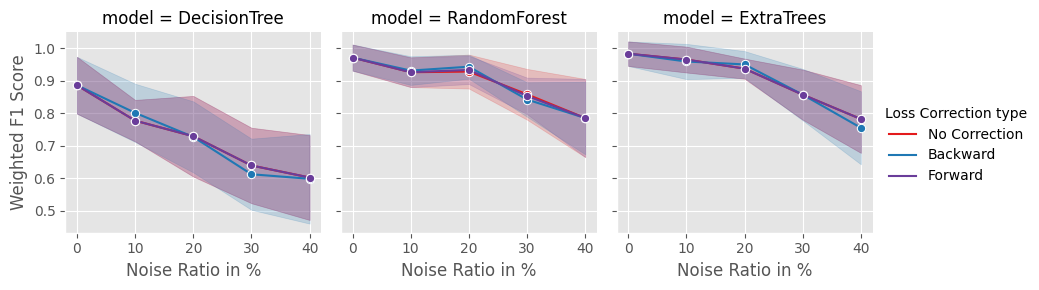}
\vspace{-5mm}
\caption{Performance of Decision Tree, Extra Trees and Random Forest models on "wine" data set. We show forward and backward loss-corrected models as well models without loss correction measured by the weighted F1 score. Reported scores are the averages of ten fold cross-validation ploted with standard deviation.}
\vspace{-1.5mm}
\label{fig:performance_plots}
\end{figure}

\begin{figure}
\centering
\includegraphics[scale=0.9]{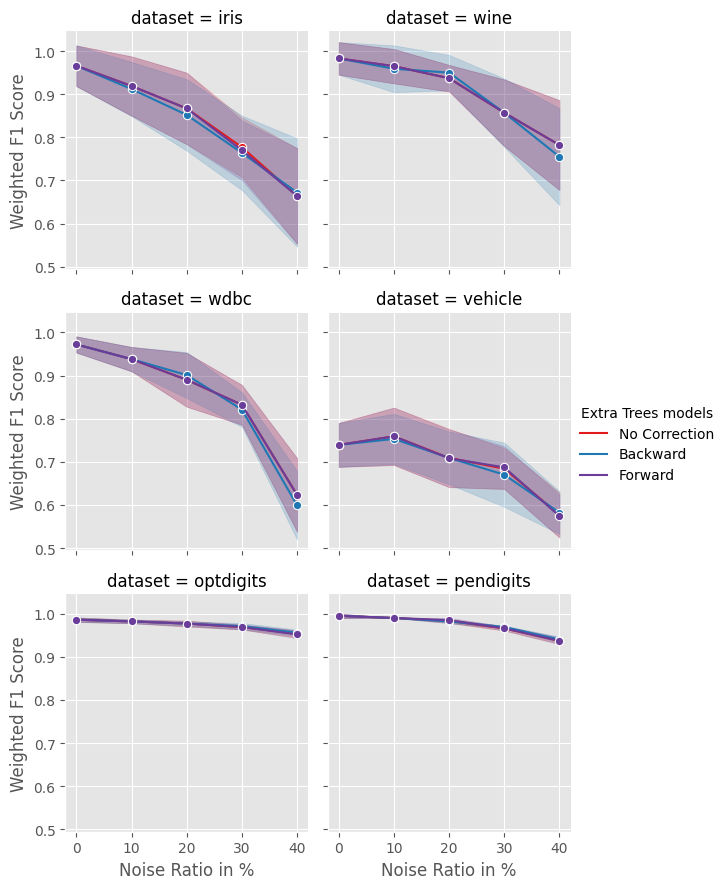}
\caption{Performance of Extra Trees forward and backward loss-corrected models as well models without loss correction measured by the weighted F1 score on six benchmarking data sets. Reported scores are the averages of ten fold cross-validation ploted with standard deviation.}
\label{fig:performance_plots_et}
\end{figure}

Figure~\ref{fig:performance_plots} shows the results on "wine" data set. The results of complete six data sets are included in Appendix Figure~\ref{fig:performance_plots_dt}, Figure~\ref{fig:performance_plots_rf}, Figure~\ref{fig:performance_plots_et}.
Overall, there is a lack of evidence that either type of loss correction improves performance of trees under label noise, confirming our theoretical findings.
\vspace{-2mm}

\textbf{CONCLUSION}
\vspace{-1.5mm}

Based on our findings, we can conclude that popular techniques of loss design employed in deep learning algorithms to robustly learn models in the presence of label noise are not applicable to decision tree induction. As a result, there is a need for alternative methods that cater specifically to robust learning of decision trees.

\pagebreak

\textbf{ACKNOWLEDGEMENTS}

This work was partially funded by the Defense Advanced Research Projects Agency under award FA8750-17-2-0130.

\textbf{URM STATEMENT}

The authors acknowledge that the paper meets the URM criteria of the ICLR 2024 Tiny Papers Track.

\bibliography{iclr2023_conference_tinypaper}
\bibliographystyle{iclr2023_conference_tinypaper}

\appendix

\section{Theory and proofs}
This section contains proofs and other support of theoretical claims made in the main paper.

\subsection{Tree fitting as loss minimization}
\label{sec:loss_equivalence}

Decision tree fitting recursively splits the input space into leaves
such that each split maximally reduces the impurity of the labels in each leaf,
summed over the leaves weighted by the number of samples in each.
Here we show that these impurity functions are
equivalent to loss functions used in parametric learning.
We will use the following identities:
\begin{enumerate}
\item The total weight of data at a given leaf is $w_\ell=\sum_{i=1}^n \mu_\ell(x_i)$.
\item A leaf value is the mean label of data belonging to the leaf: $v_\ell=\frac{1}{w_\ell}\sum_{i=1}^n \mu_\ell(x_i)y_i$.
\item The predicted value is the value of the leaf to which the sample belongs: $\hat y_i = \sum_\ell \mu_\ell(x_i) v_\ell$.
\end{enumerate}

First we show that Gini gain is equivalent to mean squared error loss.
The impurity associated with Gini gain is as follows.
\begin{align*}
&\frac{1}{n}\sum_\ell w_\ell (1- v_\ell^\top v_\ell)  \\
&\frac{1}{n}\sum_\ell w_\ell (1- 2v_\ell^\top v_\ell+v_\ell^\top v_\ell)  \\
&=1-\frac{2}{n}\sum_\ell w_\ell v_\ell^\top v_\ell +\frac{1}{n}\sum_\ell w_\ell v_\ell^\top v_\ell \\
&=1-\frac{2}{n}\sum_\ell w_\ell v_\ell^\top\left(\frac{1}{w_\ell}\sum_{i=1}^n \mu_\ell(x_i)y_i\right) 
+\frac{1}{n}\sum_\ell\left(\sum_{i=1}^n \mu_\ell(x_i)\right)v_\ell^\top v_\ell \\
&=1-\frac{2}{n}\sum_\ell v_\ell^\top\left(\sum_{i=1}^n \mu_\ell(x_i)y_i\right)
+\frac{1}{n}\sum_{i=1}^n\sum_\ell\mu_\ell(x_i)v_\ell^\top v_\ell \\
\intertext{For the rightmost term, recall that, 
for each $i$, $\mu_\ell(x_i)$ is 1 for exactly one leaf and zero for all others.}
&=1-\frac{2}{n}\sum_{i=1}^n \left(\sum_\ell \mu_\ell(x_i)v_\ell\right)^\top y_i
+\frac{1}{n}\left(\sum_{i=1}^n\sum_\ell\mu_\ell(x_i)v_\ell\right)^\top \left(\sum_{i=1}^n\sum_\ell\mu_\ell(x_i)v_\ell\right) \\
&=\frac{1}{n}\sum_{i=1}^n 1-2\hat y^\top y_i+\hat y^\top \hat y \\
\intertext{Since $y_i$ is an indicator, $y_i^\top y_i=1$.}
&=\frac{1}{n}\sum_{i=1}^n y_i^\top y_i+\hat y^\top \hat y-2 y^\top \hat y_i \\
&=\frac{1}{n}\sum_{i=1}^n \lVert y - \hat y\rVert^2 
\end{align*}
This is the mean squared error.

Next we show that information gain is equivalent to cross-entropy loss.
The impurity associated with information gain is as follows.
\begin{align*}
&\frac{1}{n}\sum_\ell w_\ell v_\ell^\top \log v_\ell  \\
&= \frac{1}{n}\sum_\ell w_\ell\frac{1}{w_\ell}\left(\sum_{i=1}^n \mu_\ell(x_i)y_i\right)^\top \log v_\ell \\
&= \frac{1}{n}\sum_\ell \sum_{i=1}^n \mu_\ell(x_i) y_i^\top \log v_\ell \\
&= \frac{1}{n}\sum_{i=1}^n  y_i^\top \left(\sum_\ell\mu_\ell(x_i) \log v_\ell \right) \\
\intertext{For each $i$, $\mu_\ell(x_i)$ is 1 for exactly one leaf and zero for all others.}
&= \frac{1}{n}\sum_{i=1}^n  y_i^\top \log \left(\sum_\ell\mu_\ell(x_i) v_\ell \right) \\
&= \frac{1}{n}\sum_{i=1}^n  y_i^\top \log \hat y
\end{align*}
This is the cross-entropy loss.

\subsection{Proof of Theorem~\ref{thm:forward_correction}}
\label{sec:proof_forward}

Assume the leaf values that minimize the loss are the weighted mean
of the training data labels in each leaf.
Let $v_\ell$ denote this weighted mean for leaf $\ell$.
Forward loss correction uses loss 
$\mathcal{L}_T(y_i,\hat y_i)=\mathcal{L}(y_i,T\hat y_i)$;
and so the minimizing leaf value $v^{(T)}_\ell$ is
\begin{align*}
v^{(T)}_\ell
&=\underset{v}{\arg\min} \sum_{i=1}^n \mu_\ell(x_i)\mathcal{L}_T(y_i,v) \\
&=\underset{v}{\arg\min} \sum_{i=1}^n \mu_\ell(x_i)\mathcal{L}(y_i,Tv).
\end{align*}
By substitution we have
\begin{align*}
Tv^{(T)}_\ell
&=\underset{v}{\arg\min} \sum_{i=1}^n \mu_\ell(x_i)\mathcal{L}(y_i,v) \\
&=v_\ell
\end{align*}
so $v_\ell^{(T)}=T^{-1}v_\ell$.
Then the total corrected loss is
\begin{equation*}
\begin{split}
&\sum_{\ell} \sum_{i=1}^n \mu_\ell(x_i)\mathcal{L}_T(y_i,v_\ell^{(T)}) \\
&= \sum_{\ell} \sum_{i=1}^n \mu_\ell(x_i)\mathcal{L}(y_i,TT^{-1}v_\ell) \\
&= \sum_{\ell} \sum_{i=1}^n \mu_\ell(x_i)\mathcal{L}(y_i,v_\ell),
\end{split}
\end{equation*}
which is the same as the loss without forward correction.

\subsection{Backward loss correction for decision trees}
\label{sec:proof_backward}

Let $\vec{\mathcal{L}}:\mathbb{R}^c\to\mathbb{R}^c$ map a prediction $\hat y$
to the loss for each possible observed class, that is,
$\vec{\mathcal{L}}(\hat y)=(\mathcal{L}(e_k,\hat y))_{k=1}^c$.
Backward loss correction uses corrected loss
$\mathcal{L}_T(y, \hat y)=y^\top T^{-\top}\vec{\mathcal{L}}(\hat y)$
where $T^{-\top}$ is the inverse transpose of $T$.

For a decision tree, the backward corrected loss is as follows.
\begin{align*}
&\sum_{i=1}^n \mathcal{L}_T(y_i, \hat y_i) \\
&=\sum_{\ell} \sum_{i=1}^n \mu_\ell(x_i) \mathcal{L}_T(y_i, v_\ell) \\
&=\sum_{\ell} \sum_{i=1}^n \mu_\ell(x_i) y_i^\top T^{-\top}\vec{\mathcal{L}}(v_\ell) \\
&=\sum_{\ell}  \sum_{k=1}^c \left(\sum_{i=1}^n \mu_\ell(x_i) y_i^\top T^{-\top}\right)_k\mathcal{L}(e_k, v_\ell) \\
&=\sum_{\ell} \sum_{k=1}^c w_k^{(T)}\mathcal{L}(e_k, v_\ell).
\end{align*}
which is the same as uncorrected loss, but with corrected weight
$$w_k^{(T)}=\left(\sum_{i=1}^n \mu_\ell(x_i) y_i^\top T^{-\top}\right)_k$$
as opposed to the uncorrected weights
$$w_k=\sum_{i=1}^n \mu_\ell(x_i)\mathbf{1}\{y_i=e_k\}.$$
Thus, assuming the minimizing value of $\mathcal{L}$ is the weighted mean,
then the minimizer of the corrected loss is likewise the weighted mean.
\begin{align*}
v_\ell^{(T)}
&=\frac{\sum_{k=1}^c w_k^{(T)}e_k}{\sum_{k=1}^c w^{(T)}_k} \\
&=\frac{\sum_{k=1}^c (w_1^{(T)},\dots,w_c^{(T)})}{\sum_{k=1}^c w^{(T)}_k} \\
&=\frac{\sum_{i=1}^n \mu_\ell(x_i) y_i^\top T^{-\top}}{\sum_{i=1}^n \mu_\ell(x_i) \sum_k(y_i T^{-\top})_k} \\
\intertext{Since $y_i^\top T^{-\top}$ represents a probability, it sums to 1.}
&=\frac{\sum_{i=1}^n \mu_\ell(x_i) y_i^\top T^{-\top}}{\sum_{i=1}^n \mu_\ell(x_i)}
\end{align*}
This is the usual leaf value, but computed with labels $y_i$ changed to 
$y_i^\top T^{-\top}$.
Moreover, plugging this back into the loss, we have the following.
\begin{align*}
\sum_{i=1}^n \mathcal{L}_T(y_i, \hat y_i)
&=\sum_{\ell} \sum_{i=1}^n \mu_\ell(x_i) y_i^\top T^{-\top}\vec{\mathcal{L}}(v^{(T)}_\ell)
\intertext{Assume $y^\top \vec{\mathcal{L}}(\hat y)=\mathcal{L}(y,\hat y)$
(shown later).}
&=\sum_{\ell} \sum_{i=1}^n \mu_\ell(x_i) \mathcal{L}(y_i^\top T^{-\top},v^{(T)}_\ell)
\end{align*}
This is the usual loss, but computed with labels $y_i$ changed to $y_i^\top T^{-\top}$.

Since both the corrected loss and corrected leaf values can be computed
by simply swapping in the corrected labels,
we can learn the corrected tree by simply fitting a tree as usual
to the corrected data set
$\{(x_i,y_i^\top T^{-\top})\}_{i=1}^n$, written equivalently as
$\{(x_i,T^{-1}y_i)\}_{i=1}^n$.

We now show that the assumption 
$y^\top \vec{\mathcal{L}}(\hat y)=\mathcal{L}(y,\hat y)$ 
holds for the loss functions corresponding to commonly used decision tree impurities.

For mean squared error, corresponding to Gini gain, we have
\begin{align*}
y^\top \vec{\mathcal{L}}(\hat y)
&=\sum_{k=1}^c y_k\lVert e_k - \hat y\rVert^2 \\
&=\sum_{k=1}^c y_k(1+\hat y^\top \hat y-2e_k^\top \hat y)
\intertext{Recall that $y$ is an indicator, 
so $\sum_{k=1}^c y_k=1$ and $y^\top y=1$.}
&=y^\top y+\hat y^\top \hat y-2y^\top \hat y \\
&=\lVert y - \hat y\rVert^2 \\
&=\mathcal{L}(y,\hat y).
\end{align*}

For cross-entropy loss, corresponding to information gain, we have
\begin{align*}
y^\top \vec{\mathcal{L}}(\hat y)
&=-\sum_{k=1}^c y_k(e_k^\top \log \hat y) \\
&=-\sum_{k=1}^c y_k\log \hat y_k \\
&=\mathcal{L}(y,\hat y).
\end{align*}

\subsection{Proof of Theorem~\ref{thm:symmetric_losses}}
\label{sec:proof_symmetric}

Assume the loss function $\mathcal{L}$ is non-negative
and symmetric, that is, 
for any $\hat{y}$, $\sum_{k=1}^c \mathcal{L}(e_k,\hat{y}) = C$.
Then loss at a single leaf $\ell$ is as follows.
\begin{align*}
&\sum_{i=1}^n \mu_\ell(x_i)\mathcal{L}(y_i,v_\ell)
\intertext{Let $w_k=\sum_i^n \mu_{\ell}(x_i) \mathbf{1}\{y_i=e_k\}$
denote the weight of class $k$ at leaf $\ell$.
Without loss of generality, assume $w_1\geq w_k$ for $k>1$,
that is, label 1 is a plurality label.}
&=\sum_{k=1}^c w_k \mathcal{L}(e_k,v_{\ell}) \\
&=w_1 \mathcal{L}(e_1,v_\ell) + \sum_{k=2}^c (w_k+(w_1-w_1)) \mathcal{L}(e_k,v_\ell)  \\
&=w_1 \sum_{k=1}^c \mathcal{L}(e_k,v_\ell) + \sum_{k=2}^c (w_1-w_k) \mathcal{L}(e_k,v_\ell)  \\
&=w_1 C + \sum_{k=2}^c (w_1-w_k) \mathcal{L}(e_k,v_\ell)
\ \text{ ($\mathcal{L}$ is symmetric)} \\
&\geq w_1 C 
\ \text{ ($\mathcal{L}$ is non-negative \& $w_1\geq w_k$)} \\
&= w_1\sum_{k=1}^c \mathcal{L}(e_k,e_1)
\ \text{ ($\mathcal{L}$ is symmetric)} \\
&\geq \sum_{k=1}^c w_k\mathcal{L}(e_k,e_1)
\ \text{ ($w_1\geq w_k$)} \\
&=\sum_{i=1}^n \mu_\ell(x_i)\mathcal{L}(y_i,e_1)
\end{align*}
This is the loss with $v_\ell=e_1$.
Therefore the leaf value $v_\ell=e_1$,
the indicator of a plurality label, is a minimizer of the loss.

\section{Data sets used in experiments}
\begin{table}[h]
    \centering
    \begin{tabular}{lrrrrr}

        data set & OpenML ID & $n$ & $m$ & $c$ \\
        \midrule
        iris & 61 & 150 & 4 & 3  \\
        optdigits & 28 & 5620 & 64 & 10  \\
        pendigits & 32 & 10992 & 16 & 10  \\
        vehicle & 54 & 846 & 18 & 4 \\
        wine & 187 & 178 & 13 & 3  \\
        wdbc & 1510 & 569 & 30 & 2  \\
        \hspace{3mm}
    \end{tabular}
    \caption{Description of the data sets used in experimental study.}
    \label{tab:datasets}
\end{table}

\section{Experimental results}

\begin{figure}
\centering
\includegraphics[scale=0.9]{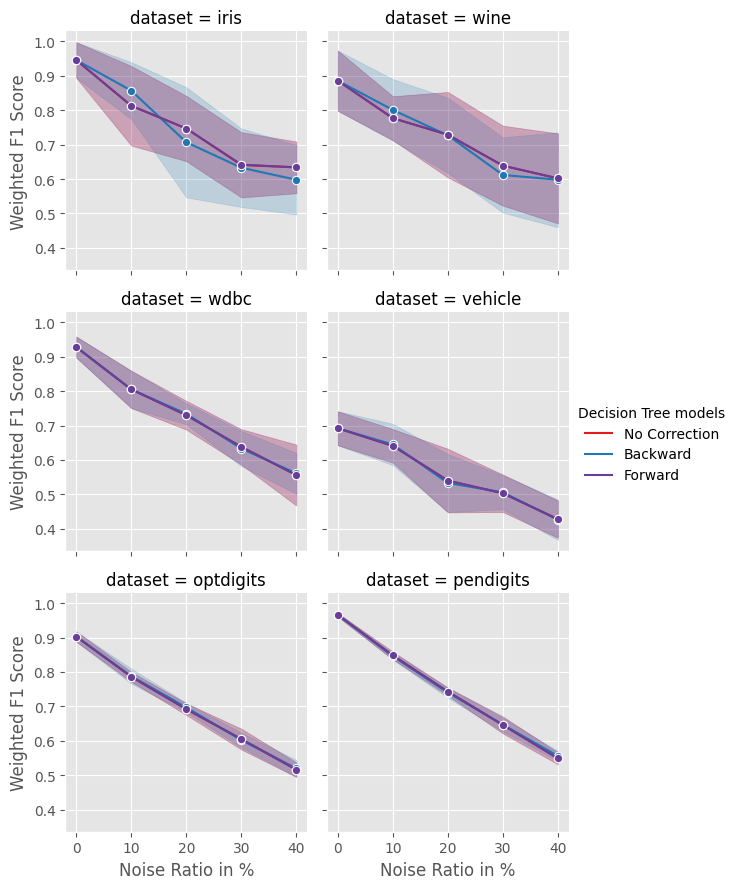}
\caption{Performance of Decision Tree forward and backward loss-corrected models as well models without loss correction measured by the weighted F1 score on six benchmarking data sets. Reported scores are the averages of ten fold cross-validation ploted with standard deviation.}
\label{fig:performance_plots_dt}
\end{figure}
\begin{figure}
\centering
\includegraphics[scale=0.9]{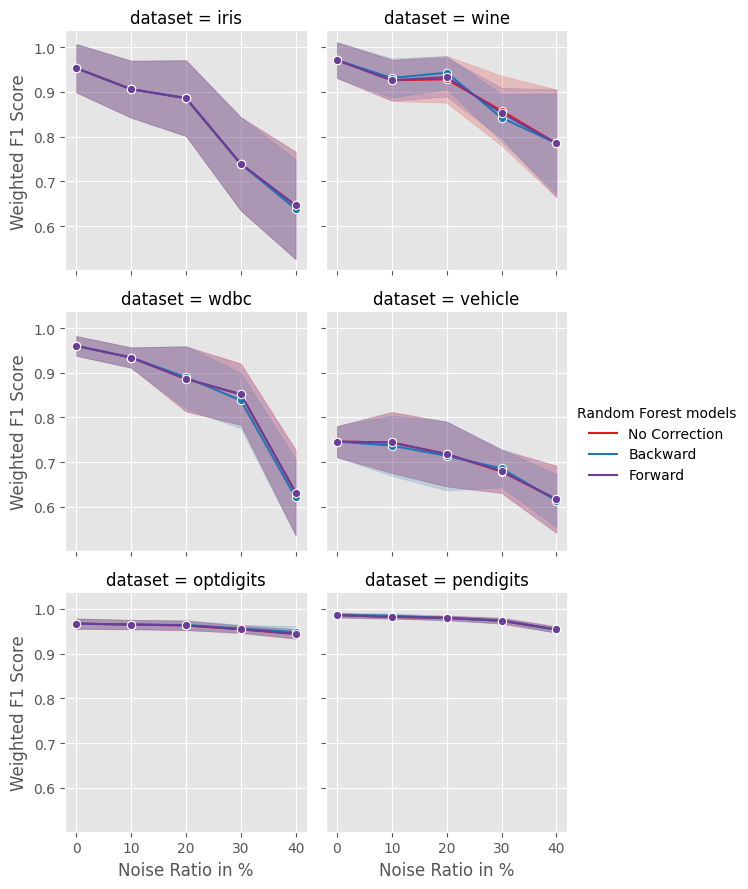}
\caption{Performance of Random Forest forward and backward loss-corrected models as well models without loss correction measured by the weighted F1 score on six benchmarking data sets. Reported scores are the averages of ten fold cross-validation ploted with standard deviation.}
\label{fig:performance_plots_rf}
\end{figure}

\begin{table}[h!]
\begin{tabular}{lllrrrrrr}
\toprule
 &  & Data set & iris & optdigits & pendigits  \\
Noise Ratio & Model & Loss Correction Type &  &  &  \\
\midrule
0 & DecisionTree & Backward & 0.946 $\pm$ 0.052 & 0.903 $\pm$ 0.015 & 0.966 $\pm$ 0.005 \\
0 & DecisionTree & Forward & 0.946 $\pm$ 0.052 & 0.903 $\pm$ 0.015 & 0.966 $\pm$ 0.005 \\
0 & DecisionTree & No Correction & 0.946 $\pm$ 0.052 & 0.903 $\pm$ 0.015 & 0.966 $\pm$ 0.005 \\
0 & ExtraTrees & Backward & 0.966 $\pm$ 0.047 & 0.986 $\pm$ 0.005 & 0.995 $\pm$ 0.005 \\
0 & ExtraTrees & Forward & 0.966 $\pm$ 0.047 & 0.986 $\pm$ 0.005 & 0.995 $\pm$ 0.005 \\
0 & ExtraTrees & No Correction & 0.966 $\pm$ 0.047 & 0.986 $\pm$ 0.005 & 0.995 $\pm$ 0.005 \\
0 & RandomForest & Backward & 0.953 $\pm$ 0.054 & 0.967 $\pm$ 0.012 & 0.986 $\pm$ 0.005 \\
0 & RandomForest & Forward & 0.953 $\pm$ 0.054 & 0.967 $\pm$ 0.012 & 0.986 $\pm$ 0.005 \\
0 & RandomForest & No Correction & 0.953 $\pm$ 0.054 & 0.967 $\pm$ 0.012 & 0.986 $\pm$ 0.005 \\
\midrule
10 & DecisionTree & Backward & 0.857 $\pm$ 0.083 & 0.788 $\pm$ 0.022 & 0.846 $\pm$ 0.010 \\
10 & DecisionTree & Forward & 0.813 $\pm$ 0.115 & 0.786 $\pm$ 0.014 & 0.848 $\pm$ 0.011 \\
10 & DecisionTree & No Correction & 0.813 $\pm$ 0.115 & 0.786 $\pm$ 0.014 & 0.848 $\pm$ 0.011 \\
10 & ExtraTrees & Backward & 0.912 $\pm$ 0.063 & 0.982 $\pm$ 0.004 & 0.990 $\pm$ 0.000 \\
10 & ExtraTrees & Forward & 0.919 $\pm$ 0.069 & 0.982 $\pm$ 0.004 & 0.990 $\pm$ 0.000 \\
10 & ExtraTrees & No Correction & 0.919 $\pm$ 0.069 & 0.982 $\pm$ 0.004 & 0.990 $\pm$ 0.000 \\
10 & RandomForest & Backward & 0.906 $\pm$ 0.064 & 0.965 $\pm$ 0.010 & 0.985 $\pm$ 0.005 \\
10 & RandomForest & Forward & 0.906 $\pm$ 0.064 & 0.965 $\pm$ 0.011 & 0.982 $\pm$ 0.004 \\
10 & RandomForest & No Correction & 0.906 $\pm$ 0.064 & 0.965 $\pm$ 0.011 & 0.982 $\pm$ 0.004 \\
\midrule
20 & DecisionTree & Backward & 0.707 $\pm$ 0.160 & 0.697 $\pm$ 0.013 & 0.740 $\pm$ 0.015 \\
20 & DecisionTree & Forward & 0.747 $\pm$ 0.095 & 0.692 $\pm$ 0.017 & 0.743 $\pm$ 0.012 \\
20 & DecisionTree & No Correction & 0.747 $\pm$ 0.095 & 0.692 $\pm$ 0.017 & 0.743 $\pm$ 0.012 \\
20 & ExtraTrees & Backward & 0.852 $\pm$ 0.083 & 0.977 $\pm$ 0.005 & 0.982 $\pm$ 0.004 \\
20 & ExtraTrees & Forward & 0.867 $\pm$ 0.083 & 0.977 $\pm$ 0.007 & 0.985 $\pm$ 0.005 \\
20 & ExtraTrees & No Correction & 0.867 $\pm$ 0.083 & 0.977 $\pm$ 0.007 & 0.983 $\pm$ 0.005 \\
20 & RandomForest & Backward & 0.886 $\pm$ 0.085 & 0.964 $\pm$ 0.011 & 0.979 $\pm$ 0.006 \\
20 & RandomForest & Forward & 0.886 $\pm$ 0.085 & 0.963 $\pm$ 0.011 & 0.980 $\pm$ 0.005 \\
20 & RandomForest & No Correction & 0.886 $\pm$ 0.085 & 0.963 $\pm$ 0.011 & 0.980 $\pm$ 0.005 \\
\midrule
30 & DecisionTree & Backward & 0.633 $\pm$ 0.114 & 0.602 $\pm$ 0.021 & 0.647 $\pm$ 0.020 \\
30 & DecisionTree & Forward & 0.642 $\pm$ 0.094 & 0.605 $\pm$ 0.030 & 0.646 $\pm$ 0.025 \\
30 & DecisionTree & No Correction & 0.641 $\pm$ 0.095 & 0.605 $\pm$ 0.030 & 0.646 $\pm$ 0.025 \\
30 & ExtraTrees & Backward & 0.764 $\pm$ 0.086 & 0.971 $\pm$ 0.007 & 0.969 $\pm$ 0.003 \\
30 & ExtraTrees & Forward & 0.770 $\pm$ 0.069 & 0.969 $\pm$ 0.006 & 0.966 $\pm$ 0.005 \\
30 & ExtraTrees & No Correction & 0.777 $\pm$ 0.069 & 0.969 $\pm$ 0.006 & 0.966 $\pm$ 0.005 \\
30 & RandomForest & Backward & 0.739 $\pm$ 0.105 & 0.956 $\pm$ 0.008 & 0.974 $\pm$ 0.005 \\
30 & RandomForest & Forward & 0.739 $\pm$ 0.105 & 0.955 $\pm$ 0.008 & 0.974 $\pm$ 0.007 \\
30 & RandomForest & No Correction & 0.739 $\pm$ 0.105 & 0.954 $\pm$ 0.008 & 0.973 $\pm$ 0.007 \\
40 & DecisionTree & Backward & 0.598 $\pm$ 0.101 & 0.520 $\pm$ 0.024 & 0.556 $\pm$ 0.014 \\
\midrule
40 & DecisionTree & Forward & 0.634 $\pm$ 0.075 & 0.516 $\pm$ 0.021 & 0.549 $\pm$ 0.017 \\
40 & DecisionTree & No Correction & 0.634 $\pm$ 0.075 & 0.516 $\pm$ 0.021 & 0.549 $\pm$ 0.017 \\
40 & ExtraTrees & Backward & 0.672 $\pm$ 0.054 & 0.956 $\pm$ 0.007 & 0.940 $\pm$ 0.007 \\
40 & ExtraTrees & Forward & 0.664 $\pm$ 0.047 & 0.952 $\pm$ 0.009 & 0.937 $\pm$ 0.008 \\
40 & ExtraTrees & No Correction & 0.664 $\pm$ 0.047 & 0.953 $\pm$ 0.008 & 0.937 $\pm$ 0.007 \\
40 & RandomForest & Backward & 0.638 $\pm$ 0.052 & 0.948 $\pm$ 0.013 & 0.953 $\pm$ 0.007 \\
40 & RandomForest & Forward & 0.646 $\pm$ 0.049 & 0.944 $\pm$ 0.011 & 0.954 $\pm$ 0.007 \\
40 & RandomForest & No Correction & 0.646 $\pm$ 0.049 & 0.944 $\pm$ 0.011 & 0.954 $\pm$ 0.007 \\
\bottomrule

\end{tabular}
\caption{Detailed performance of forward, backward loss-corrected models and models without loss correction measured by the weighted F1 score on "iris", "optdigits" and "pendigits" data sets. Scores are the averages of ten fold cross-validation reported with standard deviation.}
\end{table}

\begin{table}[h!]
\begin{tabular}{lllrrrrrr}
\toprule
 &  & Data set & vehicle & wdbc & wine  \\
Noise Ratio & Model & Loss Correction Type &  &  &  \\
\midrule
0 & DecisionTree & Backward & 0.692 $\pm$ 0.049 & 0.928 $\pm$ 0.030 & 0.886 $\pm$ 0.087 \\
0 & DecisionTree & Forward & 0.692 $\pm$ 0.049 & 0.928 $\pm$ 0.030 & 0.886 $\pm$ 0.087 \\
0 & DecisionTree & No Correction & 0.692 $\pm$ 0.049 & 0.928 $\pm$ 0.030 & 0.886 $\pm$ 0.087 \\
0 & ExtraTrees & Backward & 0.739 $\pm$ 0.051 & 0.972 $\pm$ 0.019 & 0.983 $\pm$ 0.038 \\
0 & ExtraTrees & Forward & 0.739 $\pm$ 0.051 & 0.972 $\pm$ 0.019 & 0.983 $\pm$ 0.038 \\
0 & ExtraTrees & No Correction & 0.739 $\pm$ 0.051 & 0.972 $\pm$ 0.019 & 0.983 $\pm$ 0.038 \\
0 & RandomForest & Backward & 0.746 $\pm$ 0.035 & 0.960 $\pm$ 0.022 & 0.971 $\pm$ 0.040 \\
0 & RandomForest & Forward & 0.746 $\pm$ 0.035 & 0.960 $\pm$ 0.022 & 0.971 $\pm$ 0.040 \\
0 & RandomForest & No Correction & 0.746 $\pm$ 0.035 & 0.960 $\pm$ 0.022 & 0.971 $\pm$ 0.040 \\
\midrule
10 & DecisionTree & Backward & 0.645 $\pm$ 0.060 & 0.805 $\pm$ 0.054 & 0.801 $\pm$ 0.090 \\
10 & DecisionTree & Forward & 0.641 $\pm$ 0.049 & 0.805 $\pm$ 0.054 & 0.777 $\pm$ 0.064 \\
10 & DecisionTree & No Correction & 0.641 $\pm$ 0.049 & 0.805 $\pm$ 0.054 & 0.777 $\pm$ 0.064 \\
10 & ExtraTrees & Backward & 0.753 $\pm$ 0.058 & 0.938 $\pm$ 0.028 & 0.959 $\pm$ 0.054 \\
10 & ExtraTrees & Forward & 0.759 $\pm$ 0.067 & 0.938 $\pm$ 0.028 & 0.965 $\pm$ 0.040 \\
10 & ExtraTrees & No Correction & 0.759 $\pm$ 0.067 & 0.938 $\pm$ 0.028 & 0.965 $\pm$ 0.040 \\
10 & RandomForest & Backward & 0.737 $\pm$ 0.068 & 0.934 $\pm$ 0.023 & 0.931 $\pm$ 0.044 \\
10 & RandomForest & Forward & 0.744 $\pm$ 0.068 & 0.934 $\pm$ 0.023 & 0.926 $\pm$ 0.046 \\
10 & RandomForest & No Correction & 0.744 $\pm$ 0.068 & 0.934 $\pm$ 0.023 & 0.926 $\pm$ 0.046 \\
\midrule
20 & DecisionTree & Backward & 0.533 $\pm$ 0.084 & 0.734 $\pm$ 0.031 & 0.727 $\pm$ 0.109 \\
20 & DecisionTree & Forward & 0.540 $\pm$ 0.092 & 0.730 $\pm$ 0.042 & 0.729 $\pm$ 0.124 \\
20 & DecisionTree & No Correction & 0.540 $\pm$ 0.092 & 0.730 $\pm$ 0.042 & 0.729 $\pm$ 0.124 \\
20 & ExtraTrees & Backward & 0.709 $\pm$ 0.061 & 0.901 $\pm$ 0.053 & 0.950 $\pm$ 0.041 \\
20 & ExtraTrees & Forward & 0.708 $\pm$ 0.067 & 0.890 $\pm$ 0.062 & 0.937 $\pm$ 0.031 \\
20 & ExtraTrees & No Correction & 0.709 $\pm$ 0.067 & 0.890 $\pm$ 0.062 & 0.937 $\pm$ 0.031 \\
20 & RandomForest & Backward & 0.714 $\pm$ 0.077 & 0.890 $\pm$ 0.070 & 0.943 $\pm$ 0.037 \\
20 & RandomForest & Forward & 0.718 $\pm$ 0.073 & 0.886 $\pm$ 0.073 & 0.933 $\pm$ 0.043 \\
20 & RandomForest & No Correction & 0.718 $\pm$ 0.073 & 0.886 $\pm$ 0.073 & 0.928 $\pm$ 0.052 \\
\midrule
30 & DecisionTree & Backward & 0.506 $\pm$ 0.050 & 0.633 $\pm$ 0.051 & 0.612 $\pm$ 0.110 \\
30 & DecisionTree & Forward & 0.503 $\pm$ 0.054 & 0.638 $\pm$ 0.051 & 0.639 $\pm$ 0.116 \\
30 & DecisionTree & No Correction & 0.503 $\pm$ 0.054 & 0.638 $\pm$ 0.051 & 0.639 $\pm$ 0.116 \\
30 & ExtraTrees & Backward & 0.670 $\pm$ 0.074 & 0.821 $\pm$ 0.040 & 0.857 $\pm$ 0.079 \\
30 & ExtraTrees & Forward & 0.687 $\pm$ 0.049 & 0.832 $\pm$ 0.046 & 0.857 $\pm$ 0.079 \\
30 & ExtraTrees & No Correction & 0.684 $\pm$ 0.048 & 0.832 $\pm$ 0.046 & 0.857 $\pm$ 0.079 \\
30 & RandomForest & Backward & 0.686 $\pm$ 0.042 & 0.838 $\pm$ 0.062 & 0.842 $\pm$ 0.053 \\
30 & RandomForest & Forward & 0.680 $\pm$ 0.049 & 0.852 $\pm$ 0.069 & 0.853 $\pm$ 0.056 \\
30 & RandomForest & No Correction & 0.679 $\pm$ 0.047 & 0.852 $\pm$ 0.069 & 0.858 $\pm$ 0.078 \\
\midrule
40 & DecisionTree & Backward & 0.426 $\pm$ 0.059 & 0.561 $\pm$ 0.060 & 0.598 $\pm$ 0.138 \\
40 & DecisionTree & Forward & 0.428 $\pm$ 0.052 & 0.556 $\pm$ 0.089 & 0.602 $\pm$ 0.089 \\
40 & DecisionTree & No Correction & 0.428 $\pm$ 0.052 & 0.556 $\pm$ 0.089 & 0.602 $\pm$ 0.089 \\
40 & ExtraTrees & Backward & 0.582 $\pm$ 0.049 & 0.599 $\pm$ 0.080 & 0.755 $\pm$ 0.112 \\
40 & ExtraTrees & Forward & 0.574 $\pm$ 0.050 & 0.623 $\pm$ 0.083 & 0.782 $\pm$ 0.120 \\
40 & ExtraTrees & No Correction & 0.577 $\pm$ 0.052 & 0.623 $\pm$ 0.083 & 0.782 $\pm$ 0.120 \\
40 & RandomForest & Backward & 0.614 $\pm$ 0.017 & 0.621 $\pm$ 0.018 & 0.785 $\pm$ 0.112 \\
40 & RandomForest & Forward & 0.617 $\pm$ 0.008 & 0.632 $\pm$ 0.010 & 0.785 $\pm$ 0.112 \\
40 & RandomForest & No Correction & 0.617 $\pm$ 0.008 & 0.632 $\pm$ 0.010 & 0.785 $\pm$ 0.112 \\
\bottomrule

\end{tabular}
\caption{Detailed performance of forward, backward loss-corrected models and models without loss correction measured by the weighted F1 score on "vehicle", "wdbc" and "wine" data sets. Scores are the averages of ten fold cross-validation reported with standard deviation.}
\end{table}

\end{document}

%% file: math_commands.tex
\usepackage{amsmath,amsfonts,bm}

\def\eqref#1{equation~\ref{#1}}

\def\1{\bm{1}}

\def\ra{{\textnormal{a}}}

\def\rx{{\textnormal{x}}}

\def\rva{{\mathbf{a}}}

\def\erva{{\textnormal{a}}}

\def\ervx{{\textnormal{x}}}

\def\rmA{{\mathbf{A}}}

\def\vmu{{\bm{\mu}}}
\def\vtheta{{\bm{\theta}}}
\def\va{{\bm{a}}}

\def\ve{{\bm{e}}}

\def\vx{{\bm{x}}}

\def\eva{{a}}

\def\mA{{\bm{A}}}

\def\mH{{\bm{H}}}
\def\mI{{\bm{I}}}
\def\mJ{{\bm{J}}}

\def\mX{{\bm{X}}}

\def\mSigma{{\bm{\Sigma}}}

\DeclareMathAlphabet{\mathsfit}{\encodingdefault}{\sfdefault}{m}{sl}
\SetMathAlphabet{\mathsfit}{bold}{\encodingdefault}{\sfdefault}{bx}{n}
\newcommand{\tens}[1]{\bm{\mathsfit{#1}}}
\def\tA{{\tens{A}}}

\def\tX{{\tens{X}}}

\def\gG{{\mathcal{G}}}

\def\sA{{\mathbb{A}}}
\def\sB{{\mathbb{B}}}

\def\sS{{\mathbb{S}}}

\def\emA{{A}}

\newcommand{\etens}[1]{\mathsfit{#1}}

\def\etA{{\etens{A}}}

\newcommand{\E}{\mathbb{E}}

\newcommand{\R}{\mathbb{R}}

\newcommand{\KL}{D_{\mathrm{KL}}}
\newcommand{\Var}{\mathrm{Var}}

\newcommand{\Cov}{\mathrm{Cov}}

\newcommand{\normltwo}{L^2}
\newcommand{\normlp}{L^p}

\newcommand{\parents}{Pa} 